\documentclass[a4paper,twoside,11pt]{article}

\usepackage[ansinew]{inputenc}
\usepackage[english]{babel}
\usepackage[dvips]{graphicx}
\usepackage{color}
\usepackage[page]{appendix}
\usepackage{amssymb}
\usepackage{amsmath}
\usepackage{amsfonts}
\usepackage{ntheorem}

\begin{document}

\newtheorem{theorem}{Theorem}
\newtheorem*{theorem3}[theorem]{Theorem 3}
\newtheorem*{theorem1}[theorem]{Theorem 1}
\newtheorem{definition}[theorem]{Definition}
\newtheorem{proposition}[theorem]{Proposition}
\newenvironment{proof}[1][Proof]{\textbf{#1.} }{\ \rule{0.5em}{0.5em}}
\newcommand{\domX}{{\mathcal X}}
\def\x{{\mathbf x}}

\title{Hyperspectral Image Classification with \\Support Vector Machines on Kernel Distribution Embeddings}

\author{
Gianni Franchi$^{ \star}$ \qquad Jes\'{u}s Angulo$^{\star}$
\qquad Dino Sejdinovi\'{c}$^{\dagger}$\\
{\small $^{\star}$ MINES ParisTech, PSL-Research University, 
	CMM-Centre de Morphologie Math\'{e}matique; France}\\
	{\small $^{\dagger}$ Department of Statistics, University of Oxford; United Kingdom}\\
}

%\date{January 2014}

\maketitle

\begin{abstract}
We propose a novel approach for pixel classification in hyperspectral images, leveraging on both the spatial and spectral information in the data. The introduced method relies on a recently proposed framework for learning on distributions -- by representing them with mean elements in reproducing kernel Hilbert spaces (RKHS) and formulating a classification algorithm therein. In particular, we associate each pixel to an empirical distribution of its neighbouring pixels, a judicious representation of which in an RKHS, in conjunction with the spectral information contained in the pixel itself, give a new explicit set of features that can be fed into a suite of standard classification techniques -- we opt for a well established framework of support vector machines (SVM). Furthermore, the computational complexity is reduced via random Fourier features formalism. We study the consistency and the convergence rates of the proposed method and the experiments demonstrate strong performance on hyperspectral data with gains in comparison to the state-of-the-art results. 
\\ \textbf{Keywords: } Hyperspectral images, pixelwise classification, kernel methods.
\end{abstract}

% ----------------------------------------------------------------
\section{Introduction}
\label{section:Intro}

Hyperspectral images consist of very high-dimensional pixel observations that allow reconstruction of the spectral profiles of objects imaged thanks to the acquisition of several hundred narrow spectral bands. The supervised classification of these pixels is a challenging task, which commonly arises in remote sensing imaging \cite{camps2014advances,gualtieri1999support,Fauvel2,fauvel2013advances}. Structure of the hyperspectral imagery is seldom studied in a comprehensive manner, with most approaches focusing either on spatial information building on tools available for normal imagery or with a focus on spectral information without a principled way to make use of both. We propose a novel approach to classification based on kernel embeddings of distributions which utilizes both the spatial and spectral information in the data. While aimed at hyperspectral imaging, the method we propose is general and can be applied to other types of data.
Kernel methods and support vector machines have been employed in the hyperspectral imaging in \cite{fang2015classification,li2013generalized} the pixel data is lifted into a potentially infinite-dimensional feature space, called reproducing kernel Hilbert space (RKHS), where linear separating hyperplanes are sought. However, spectral information contained in pixels is often not sufficient for such task, and as we will see, including the local / spatial information available in the imagery is key to obtain good classification accuracy. Our approach is to encode the spatial neighbourhood of each pixel as a random sample from a distribution associated to that pixel and to treat such distribution as an additional feature for classification. In order to add consistent spatial information, we also use the Hadamard multiplication of two kernels. Where one kernel is the kernel embeddings of distributions, and the other one is the linear kernel of spatial information similarly to \cite{campsValls1,fauvel2007spectral}.

In Section \ref{sec:related_work}, related work is reviewed. Section \ref{sec:background} provides the background on kernel embeddings of distributions, random features for fast approximations to kernel methods, and on mathematical morphology, which allow us to analyse and understand the geometrical structures of images. Section \ref{sec:theory} studies the consistency and convergence rate of the proposed method and experiments are given in Section \ref{sec:experiments}.

\section{Related work}
\label{sec:related_work}

Many techniques aim to include the spatial information in the classification process. Of particular interest are those combining feature space representations describing the spatial information with those describing the pixels. Morphological feature spaces have been considered in several publications, with impressive results 
\cite{Dalla_Mura1,Fauvel2,Pesaresi,lefevre}. On the other hand, kernel methods have also been studied extensively, and more particularly the compositions of kernels \cite{li2013generalized,li2015multiple,campsValls1,Fauvel1}, which allow building new feature space representations. 

We marry these approaches with a framework of \cite{Smola2007, muandet2012learning,szabo2014two,oliva2015deep}, where instead of the usual feature map, sending each data point to the feature space, a whole distribution can be represented in the RKHS. This yields a framework for learning on distributions via their representations in this RKHS. In our approach, each pixel is associated to a distribution of its neighbours -- effectively, a hyperspectral image is treated as a set of such distributions. This is similar to the approach to regression applied in \cite{szabo2014two} to the multispectral imaging data. However, the authors of \cite{szabo2014two} partition a multispectral image and classify the partitions - with a goal to obtain responses at the level of the \emph{groups of neighbouring pixels}, which suffices when the goal is to predict an averaged quantity of an image area (e.g. aerosol concentration as studied in \cite{szabo2014two}) and the pixel-level classification is not considered. Another related line of work is that of \cite{Volpi}, where they used the mean map on hyperspectral to perform a dimensionality reduction.

%DS: removed this - this is a related work section so this should not be here - I think we have already covered all this in introduction anyway!
%We have several innovations on this article. The first one is the use of the kernel mean map support vector machines (SVM), on multivariate images. So we establish a link between the average filter in a RKHS, with the SVM. Secondly, we used the product of the mean map kernel and of another spatial kernel, to build new spatial descriptors.

\section{Background}
\label{sec:background}

\subsection{Mean Map Kernel}

Let $ k : \domX \times \domX \rightarrow \mathbb{R}$ be a positive definite kernel. By Moore-Aronszajn theorem \cite{BerTho04}, there is a unique RKHS $\mathcal{H}$ of real-valued functions on $\domX$ where $\langle g,k(\cdot,x)\rangle_\mathcal{H}=g(x)$, for all $g\in\mathcal H, x\in\domX$, implying that $k$ corresponds to an inner product between features and, in particular, $k(x,x')=\langle k(\cdot,x),k(\cdot,x')\rangle_\mathcal{H}$. This means that $k(\cdot,x)$ can be viewed as a feature of $x\in \mathcal X$. For many typical choices of kernels $k$, the RKHS $\mathcal{H}$ is infinite-dimensional. Now, let $X$ denote a random variable following a distribution $\mathcal{P}$. The \emph{mean map} or the \emph{kernel embedding} \cite{Smola2007,SriGreFukLanSch10} of $\mathcal{P}$ is defined as:
\begin{eqnarray}
\mu_{\mathcal{P}} := \mathbb{E}_X [k(\cdot, X) ] = \int_\domX k(\cdot,x) \ \mathrm{d}\mathcal{P}(x), 
\end{eqnarray}
where the expectation is over $\mathcal{H}$. For \emph{characteristic} kernels \cite{Sriperumbudur2011}, which include Gaussian RBF, Matern family and many others, this embedding is injective on the space of all probability distributions (i.e. captures information on all moments, akin to a characteristic function).
Further, if we are given two random variables, $X$ following the distribution $\mathcal{P}$, and $Y$ following the distribution $\mathcal{Q}$, the inner product between the corresponding embeddings is given as
\small
\begin{eqnarray}
\langle \mu_{\mathcal{P}},\mu_{\mathcal{Q}} \rangle_\mathcal{H} = \mathbb{E}_{X,Y} [k(X, Y) ],
\end{eqnarray}
\normalsize
which is sometimes referred to as a \emph{mean map kernel}.
For a random sample $\{ x_1,\ldots,x_n \}$, drawn independently and identically distributed from $\mathcal{P}$,  we can define the empirical mean map:
\small
\begin{eqnarray}
\widehat{\mu}_{\mathcal{P}} = \frac{1}{n} \sum_{i=1}^n k(\cdot,x_i), 
\end{eqnarray}
\normalsize
and for random samples $\{ x_1,\ldots,x_n \}$ from $\mathcal{P}$ and $\{ y_1,\ldots,y_m \}$ from $\mathcal{Q}$,  we obtain the empirical mean map kernel:
\small
\begin{eqnarray}
\langle\widehat{\mu}_{\mathcal{P}},\widehat{\mu}_{\mathcal{Q}}\rangle_\mathcal{H} = \frac{1}{nm} \sum_{i=1}^n \sum_{j=1}^m k(x_i ,y_j). 
\end{eqnarray}
\normalsize
\subsection{Random features for kernels}

The computational and storage requirements for kernel methods on large datasets can be prohibitive in practice due to the need to compute and store the kernel matrix. If we consider a dataset of $n$ $D$-dimensional observations, the storage requirements are $O(n^2)$ and the calculation takes $O(Dn^2)$ operations. A remedy developed by \cite{Rahimi2007} is to approximate translation-invariant kernels in an unbiased way using a random feature representation. Namely, any translation-invariant positive definite kernel $k$, such that $ \forall(x,y) \in \mathcal{X}^2$, $k(x,y) =  \kappa(x-y)$  can be written as
$k(x,y) = \textbf{E}_{\omega\sim\Lambda}\left[\cos(\omega^\top x)\cos(\omega^\top y)\right.$ $+\left.\sin(\omega^\top x)\sin(\omega^\top y)\right]$,
where $\omega \in \mathbb{R}^D$ follows some distribution $\Lambda$ (spectral measure of the kernel). Thus, by sampling i.i.d. vectors $\omega_1,\ldots,\omega_N$ from $\Lambda$, we can approximate kernel  $k$  by $\hat{k}$ defined by:
$\hat{k}(x,y) = \frac{1}{N}\sum_{j=1}^N \left(\cos(\omega_j^\top x)\cos(\omega_j^\top y)\right.$ $+\left. \sin(\omega_j^\top x)\sin(\omega_j^\top y)\right)$,
so that the original feature map $k(\cdot, x)$, potentially living in an infinite-dimensional space, is approximated by an explicit $2N-$dimensional feature vector:
\footnotesize
\begin{eqnarray}	
\hat{Z}(x) = \sqrt{\frac{1}{N}}\begin{bmatrix}\cos(\omega_1^\top x), \ldots , \cos(\omega_N^\top x), \sin(\omega_1^\top x),\ldots, \sin(\omega_N^\top x)\end{bmatrix}^T.
\end{eqnarray}
\normalsize
Thus, the mean map and the mean map kernel can be estimated using these finite-dimensional representations. In this contribution, we will focus on Gaussian RBF kernels for which the spectral measure $\Lambda$ is also Gaussian.

\subsection{Random features mean map on hyperspectral images}
Let us now turn our attention to a hyperspectral image $h$. Around each pixel location $x_i$, we consider a square patch $\mathcal{P}_{x_i}^{(s)}$ of size $s$ where we will treat the pixels as a random sample from a distribution $\mathcal{P}_i$ specific to the location $x_i$. Instead of calculating the kernel between individual data points, we will calculate kernel between these distributions.
An empirical mean map kernel is thus given simply by:
\begin{eqnarray}
K_{mm}(x_i,x_j)&= &\langle \hat \mu_{P_i}, \hat \mu_{P_j} \rangle_\mathcal H \\\nonumber
&=&\frac{1}{s^2} \sum_{l_1 \in \mathcal{P}_{x_i}}  \sum_{l_2  \in \mathcal{P}_{x_j}}  k(h(x_{l_1}),h(x_{l_2}))\\\nonumber
&\approx&\frac{1}{s^2} \sum_{l_1 \in \mathcal{P}_{x_i}}  \sum_{l_2  \in \mathcal{P}_{x_j}}  \hat{Z}(h(x_{l_1}))^\top\hat{Z}(h(x_{l_2})),
\end{eqnarray}
where $h(x)$ denotes the measurement vector at location $x$ and in the last line we employ a random feature approximation of $k$.

It should be noted that there may be outliers in a patch, which can damage the estimation of the mean. Similarly to the work of \cite{flaxman}, we proposed to use a weighted mean map, where the weights depend on spatial information. The kernels we obtain, called convolutional kernels, have also been used in \cite{mairal2014}. In contrast to \cite{mairal2014}, however, we will use random feature expansions to explicitly represent the feature space.

The convolutional kernel is defined as:
\small
\begin{eqnarray*}
\widehat{KCN}(x_i,x_j)=  \sum_{l1 \in \mathcal{P}_{x_i}}  \sum_{l2  \in \mathcal{P}_{x_j}}  \| h(x_{l1}) \|_2  \| h(x_{l2})\|_2 \\ e^{-\frac{1}{2\beta^2} \| x_{l1}-x_{l2} \|_2 } e^{-\frac{1}{2\sigma^2}\| \tilde{h}(x_{l1})-\tilde{h}(x_{l2}) \|_2 },
\end{eqnarray*}
\normalsize
where $\tilde{h}$ represents a normalised version of $h$, such that for all $i \in [1,n]$ =$\| \tilde h(x_i)\|_2 =1$. So we do a product of a kernel on the positions, another on the magnitudes, and a third one is an RBF kernel between spectra.

This formula can be interpreted as a weighted mean map which is defined by:
\scriptsize
\begin{eqnarray}\label{KCN1}
\widehat{KCN}(x_i,x_j)= \langle\widehat{\mu}_{P(\mathcal{P}_{x_i})},\widehat{\mu}_{P(\mathcal{P}_{x_j})}\rangle  \\
\widehat{KCN}(x_i,x_j)= \frac{1}{s^2}  \sum_{l1 \in \mathcal{P}_{x_i}}  \sum_{l2  \in \mathcal{P}_{x_j}} \| h(x_{l1}) \|_2 . \| h(x_{l2})\|_2 \hat{\mathbf{k}}(x_{l1},x_{l2}),
\end{eqnarray}
\normalsize
where $\hat{\mathbf{k}}$ is a positive definite kernel arising from the random feature space expansion.

\section{Theoretical Analysis}
\label{sec:theory}
%This section would be dedicated to do comments on \cite{szabo2014two,muandet2012learning,cortes2013learning}. The authors of   \cite{szabo2014two} proposed to derived concentration inequalities to show how regression of an estimated mean map can be consistent. On \cite{muandet2012learning},  I was thinking that the authors proposed to calculate the difference between the SVM on mean map, and the SVM on all the data of the different sets, considering the expected risk function. However after our meeting I understood that I was wrong, so I will do it. Then on \cite{cortes2013learning}, the authors proposed new inequalities for the SVM between the expected risk function and the empirical risk function.
%Based on the previous articles, we can proposed new inequalities for the SVM mean map, I follow the notation of \cite{szabo2014two}.

Let us consider that the data are partitioned into sets following the same distribution, then the structure of our data is given by $ \{(\{x_{i,n} \}_{n=1}^{N_i} ,y_i)\}_{i=1}^l $ with $x_{i,1}, \ldots, x_{i,N_i} \overset{i.i.d.}{\sim} x_i$, where $(x_i,y_i)$ are drawn from a joint meta distribution $\mathcal{M}$. We follow the notation of \cite{szabo2014two}. Let us denote $\Phi$ a loss function. Let us write the following expected risk function of the data for the SVM problem:
\small
\begin{eqnarray}
\mathcal{R}(f)=  \inf_{f \in \mathcal{H}} \mathbb{E}_{(x,y)\sim\mathcal{M}} \left( \Phi(f(x)y)\right)
\end{eqnarray}
\normalsize
We can modify it to mean map embedding classification problem :
\small
\begin{eqnarray}
\mathcal{R}_{\mu}(f)=  \inf_{f \in \mathcal{H}} \mathbb{E}_{(x,y)\sim\mathcal{M}} \left( \Phi(f(\mu_x)y)\right)
\end{eqnarray} 
\normalsize
We can also write the empirical risk function, for mean map embedding classification problem :
\small
\begin{eqnarray}
 \hat{\mathcal{R}}_{\mu}(f)=  \inf_{f \in \mathcal{H}} \frac{1}{n}\sum_{i=1}^n \left( \Phi(f(\mu_{x_i})y_i) \right)
\end{eqnarray}
\normalsize
Finally we can also write the empirical risk function, for the empirical mean map embedding classification problem :
\small
\begin{eqnarray}
 \hat{\mathcal{R}}_{\hat{\mu}}(f)=  \inf_{f\in \mathcal{H}} \frac{1}{n}\sum_{i=1}^n \left( \Phi(f(\hat{\mu}_{x_i})y_i) \right)
\end{eqnarray}
\normalsize  
Then we would like to obtain an inequality between $\mathcal{R}_{\mu}(f)$ and $ \hat{\mathcal{R}}_{\hat{\mu}}(f)$. To do that, inspired by \cite{muandet2012learning}, we derive a inequality $\mathcal{R}_{\mu}(f)$ and $ \mathcal{R}(f)$ : \\
 
\begin{theorem}
 
 Given that $x \sim P$ an arbitrary probability distribution with variance $\sigma^2$, a Lipschitz continuous function $f :  \mathbb{R} \rightarrow \mathbb{R}$ with constant $C_f$, an arbitrary loss function $\Phi:  \mathbb{R} \rightarrow \mathbb{R}$ that is Lipschitz continuous in the second argument with constant $C_l$ , it follows that :
\begin{eqnarray}
  \mathcal{R}_{\mu}(f) - \mathcal{R}(f)  \leq C_l C_f^2 \mathbb{E}_{(x)}  \| x-\mu_{x} \|^2 \mathbb{E}_{(y)} ( y^2)
\end{eqnarray}

 \end{theorem}
The proof of this theorem can be found on the supplementary materials.
Then we might use \cite{cortes2013learning} where we have an inequality between $\mathcal{R}_{\mu}(f)$ and $ \hat{\mathcal{R}}_{\mu}(f)$ : 

\begin{theorem}
 
Let $ \mathcal{G} = \Phi(\mathcal{H}, .)$ denote the loss class, let $ \mathcal{R}_n(\mathcal{G} )$ denote the Rademacher complexity. Let $ \Sigma(\mathcal{G})^2 = \sup_{g \in \mathcal{G}} \mathbb{E}(g^2)$ be a bound on the variance of the functions in
$\mathcal{G}$. If the trace of the kernel is bounded, the loss function $\Phi:  \mathbb{R}  \rightarrow \mathbb{R}$ that is Lipschitz continuous, for any $\delta>0$ , the following bound holds with probability at least $1-\delta$
\scriptsize
\begin{eqnarray*}
 \hat{\mathcal{R}}_{\mu}(f)- \mathcal{R}_{\mu}(f)  \leq 8 \mathcal{R}_n(\mathcal{G} ) + \Sigma(\mathcal{G}) \sqrt{\frac{8 \log(2/\delta)}{n}}+ \frac{3 \log(2/\delta)}{n}
\end{eqnarray*}
\normalsize
 \end{theorem}
\begin{theorem}
 
 Given that $f :  \mathbb{R} \rightarrow \mathbb{R}$ is a Lipschitz continuous function  with constant $C_f$, an arbitrary loss function $\Phi:  \mathbb{R} \rightarrow \mathbb{R}$ that is Lipschitz continuous with constant $C_{l2}$ , it follows that :
\small
\begin{eqnarray*}
  \hat{\mathcal{R}}_{\hat{\mu}}(f) -\hat{\mathcal{R}}_{\mu}(f)  \leq  \frac{1}{n}C_{l}C_f^2 \hat{\mathbb{E}}\left( \| \mu_x -\hat{\mu}_x \|^2\right)\hat{\mathbb{E}}\left( (y)^2\right)
\end{eqnarray*}
\normalsize
 \end{theorem}
The proof of this theorem can be found on the supplementary materials.

 We also need the following theorem proved in \cite{Smola2007}

\begin{theorem}
 Assume that $\|g\|_{\infty} \leq R$ for all $g \in \mathcal{H}$
  with  $\|g\|_{\mathcal{H}} \leq 1$, and that $k$ is an universal kernel. Then with probability at least $1 - \delta $ :
 
\begin{eqnarray*}
  |\mu[P] - \mu[X]| \leq 2\mathcal{R}_n(\mathcal{H}, P) + R\sqrt{\log\left(1/\delta\right)/n}
\end{eqnarray*}
\normalsize
where $\mathcal{R}_n(\mathcal{H}, P)$ denotes the Rademacher average associated with $P$ and
$\mathcal{H}$.
 \end{theorem}

Then by combining the previous theorems we easily have the following theorem.
\begin{theorem}\label{theorem1}
Given the conditions of the previous theorems. Then with probability at least $1 - \delta $ :
 \scriptsize
\begin{eqnarray*}
  \hat{\mathcal{R}}_{\hat{\mu}}(f) - \mathcal{R}(f) \leq C_l C_f^2 \left[ \mathbb{E}_{(x)}  \| x-\mu_{x} \|^2 \mathbb{E}_{(y)} ( y^2)\right. \\
   \left. + \left( 2\mathcal{R}_n(\mathcal{H}, P) + R\sqrt{\log\left(1/\delta\right)/n} \right)\hat{\mathbb{E}}\left( (y)^2\right) \right] \\
+8 \mathcal{R}_n(\mathcal{G} ) + \Sigma(\mathcal{G}) \sqrt{\frac{8 \log(2/\delta)}{n}}+ \frac{3 \log(2/\delta)}{n}
\end{eqnarray*}
\normalsize
 \end{theorem}

\section{Experiments}
\label{sec:experiments}

We evaluate the classification accuracy of the proposed approach using two standard datasets: the AVIRIS Indian Pines, and the ROSIS
University of Pavia. The first data set is an image of dimension $145 \times 145$ pixels, with $D = 224$ spectral bands and its geometrical
resolution is of 3.7 m. The training set is composed of $80$ pixels, and the image is composed of $16$ classes. The dimensions of the second data set are $ 610 \times 340$ pixels, with $D = 103$ spectral bands and its geometrical resolution is of 1.3 m.  The training set is composed of $3921$ pixels and a testing set of $42776$ pixels, and the image is composed of $9$ classes. There is a commonly used testing set for Pavia data, and we report performance on this testing set. In the first data set there is no testing set so we generate 20 Monte-Carlo simulations, selecting randomly 5 pixels per class, then aggregate the result of the classification. We used the Morphological Profile (MP) feature \cite{Pesaresi, Dalla_Mura1} space which is commonly used in pixel classifiation and is described in the supplementary material. We also use the product of the two kernels where one is the MP kernel and the other is the KMM kernels. This kind of techinique has been previously explored in \cite{li2013generalized,li2015multiple,campsValls1}. In contrast to the previous work, which approximate this product of kernels thanks to addition of kernels, we can do the real multiplication since we work with finite dimension Hilbert spaces. The results of classification are reported in Table \ref{table1} (Indian Pine),  and Table \ref{table2} (Pavia). The classification algorithm used is the C-SVM \cite{CC01a} were the parameter $C$ was selected with $5-$fold cross-validation on a grid $C=2^i$,with $i \in [-15,15]$. The results on Table \ref{table1} and \ref{table2} show us that kernel mean map can perform as well as state of art results on these images. The size of scale $s$ seems to be important, this is linked with the theorem \ref{theorem1}, where we see that increasing the size of the scale increase in a way the size of the training set. An explanation of the parameters of evaluation can be found in \cite{fauvel2007spectral} ( pages 166-167) and also in the supplementary materials.
%dino: commented below, since already said this
%To evaluate the perfomance of our algorithm we use has morphological feature which are explained in the supplementary materials, and also in [ref].

\begin{table}[width=0.35\columnwidth]
\begin{center}
\tiny
\begin{tabular}{l l l l l}
kernel & \bf parameters   & \bf OA & \bf kappa statistic  & \bf AA  \\

\hline \hline
linear kernel  & ~   & $54.6 \pm 3.3$      & $49.5 \pm 3.5$ & $58.2 \pm 2.7$\\
\hline
 random rbf    & ~    & $53.9 \pm 2.7$ &  $48.6 \pm 3.0$ & $58.0 \pm 2.8$\\
\hline
 $\hat{Kmm}$    & s=3    &$57.6 \pm 4.2$    & $48.1 \pm 4.3$ & $59.1 \pm 3.8$\\
\hline
 $\hat{Kmm}$    & s=10    &$66.5 \pm 3.2$    & $62.0 \pm 3.3$ & $65.4 \pm 3.8$\\
\hline
 $\hat{Kmm}$    & s=15    &$70.0 \pm 4.1$    & $66.4 \pm 3.9$ & $68.6 \pm 2.3$\\
\hline
 $\hat{Kmm}$    & s=20    &$70.15 \pm 3.5$    & $66.7 \pm 3.8$ & $64.1 \pm 2.3$\\
\hline
 $K_{MP}$    & ~    &$62.9 \pm 4.6$    & $58.5 \pm 5.5$ & $66.5 \pm 2.3$\\
\hline
 $K_{MP} \times \hat{Kmm}$    & s=15    & $73.0 \pm 3.7$    & $69.7 \pm 3.7$ & $76.3 \pm 2.3$\\
\hline
 $\widehat{KCN}$    & s=7    &  $\bf 77.9 \pm 3.4$    & $ \bf 76.4 \pm 3.7$ & $\bf 74.8 \pm 2.4$\\
\hline
 $\widehat{KCN}$    & s=15    & $73.0 \pm 3.7$    & $69.7 \pm 3.7$ & $76.3 \pm 2.3$\\
\hline
\hline
\end{tabular}
\end{center}
\normalsize \caption{Overall accuracy, kappa statistic, and Average accuracy obtained for different kernels, applied on the  AVIRIS Indian Pines hyperspectral data set. I have run 20 Monte Carlo simulations. I have selected  on the training set just 5 samples per class.} \label{table1}
\end{table}

\begin{table}[width=0.35\columnwidth]
\begin{center}
\tiny
\begin{tabular}{l l l l l}
kernel & \bf parameters   & \bf OA & \bf kappa statistic  & \bf AA  \\

\hline \hline
linear kernel  & ~   & $73.2$      & $66.6$ & $78.5$\\
\hline
 random rbf    & ~    & $78.1 \pm 2.5$ &  $74.5 \pm 2.5$ & $80.4 \pm 1.8$\\
\hline
 $\hat{Kmm}$    & s=3    &$90.0 \pm 2.1$    & $87.4 \pm 2.5$ & $89.7 \pm 1.1$\\
\hline
 $\hat{Kmm}$    & s=10    &$93.2 \pm 1.4$    & $91.1 \pm 1.7$ & $93.2 \pm 0.8$\\
\hline
 $\hat{Kmm}$    & s=15    &$93.9 \pm 0.8$    & $91.9 \pm 1.0$ & $93.4 \pm 0.5$\\
\hline
 $\hat{Kmm}$    & s=20    &$87.5 \pm 2.6$    & $83.0 \pm 2.5$ & $88.0 \pm 1.2$\\
\hline
 $K_{MP}$    & ~    &$97.1$    & $96.2$ & $96.7$\\
\hline
 $K_{MP} \times \hat{Kmm}$ & s=10      &  $\bf97.4 \pm 0.6$    & $\bf 96.4 \pm 0.7$ & $\bf97.3 \pm 0.6$\\
\hline
 $\widehat{KCN}$    & s=10    &  $96.2 \pm 0.9$    & $94.9 \pm 1.2$ & $94.0 \pm 1.5$\\
\hline
 $\widehat{KCN}$    & s=15    &  $96.4 \pm 0.8$    & $95.1 \pm 1.2$ & $94.4 \pm 1.8$\\
\hline
\hline
\end{tabular}
\end{center}
\normalsize \caption{Overall accuracy, kappa statistic, and Average accuracy obtained for different kernels, applied on the  University of Pavia hyperspectral data set. We used the classical training set.} \label{table2}
\end{table}

\normalsize

\begin{figure}[h]
\begin{center}
\begin{tabular}{ccc}
\includegraphics[width=0.23\columnwidth]{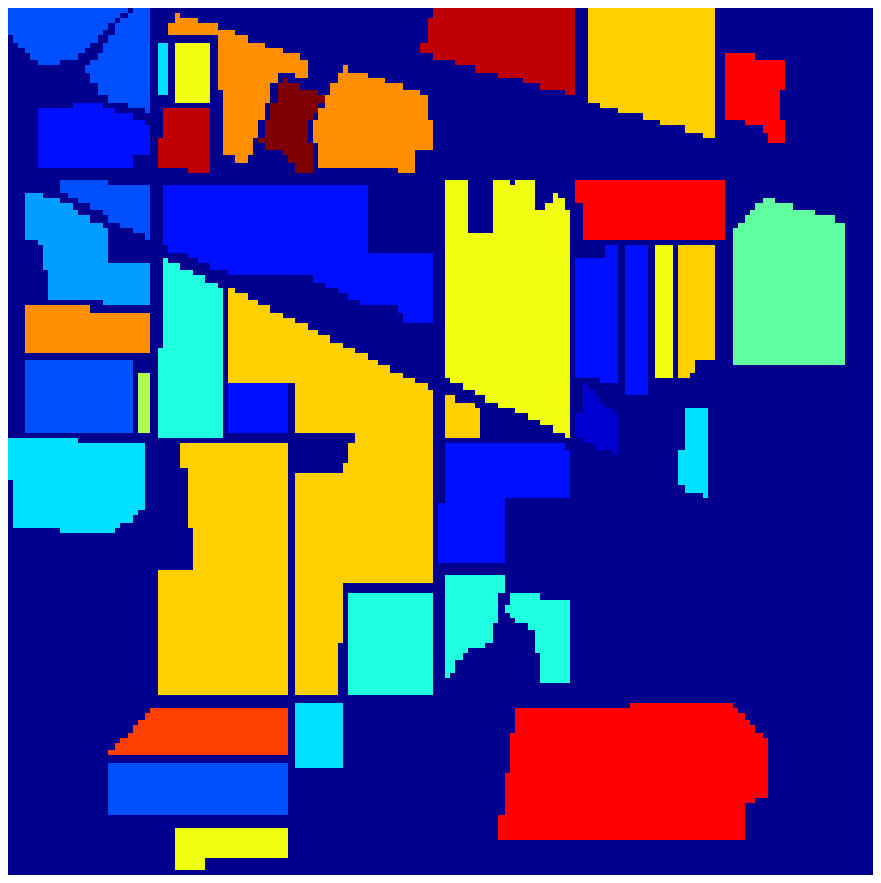}&
\includegraphics[width=0.23\columnwidth]{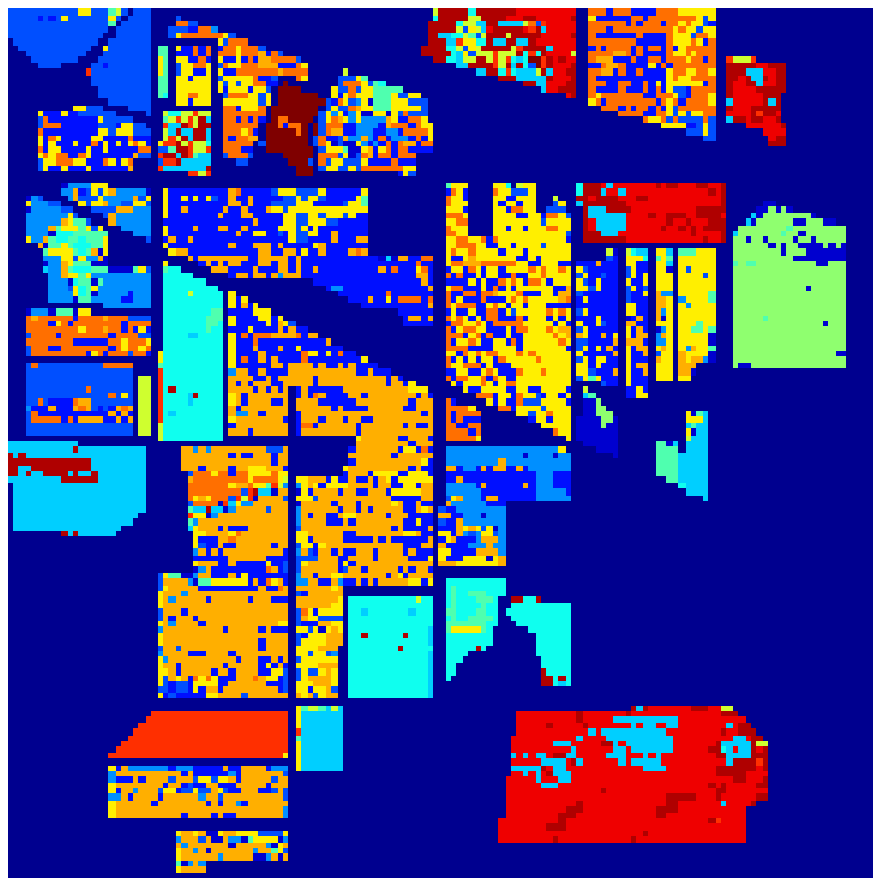}&
\includegraphics[width=0.23\columnwidth]{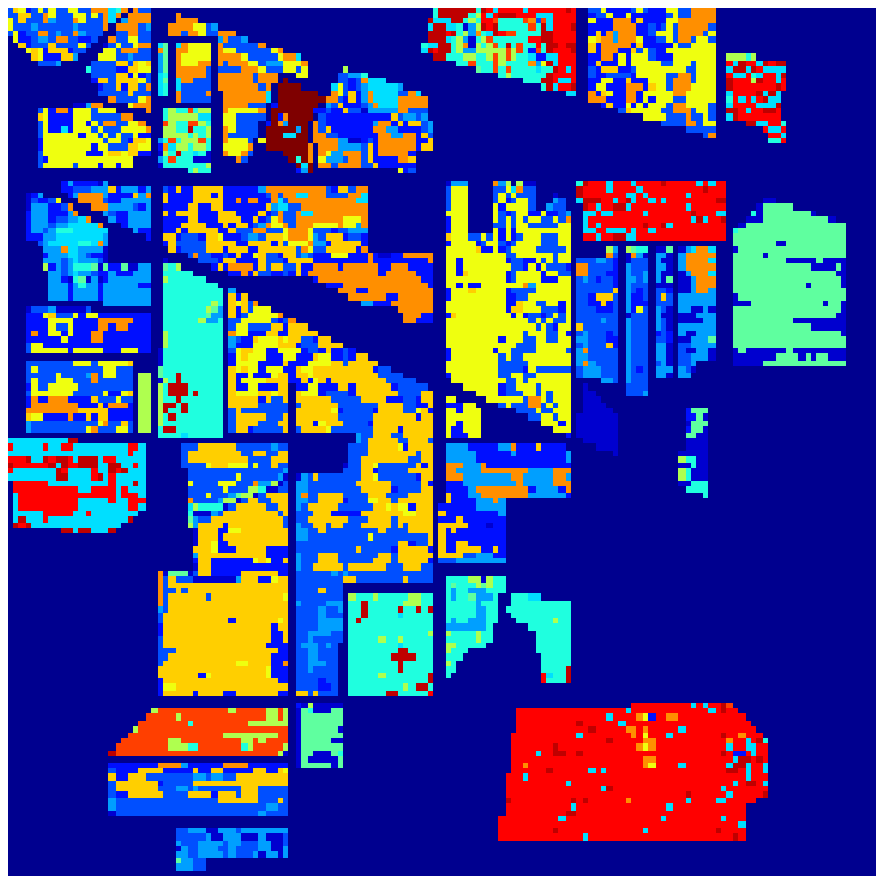}\\
{\small (a)} & {\small (b)} & {\small (c)}\\
\includegraphics[width=0.23\columnwidth]{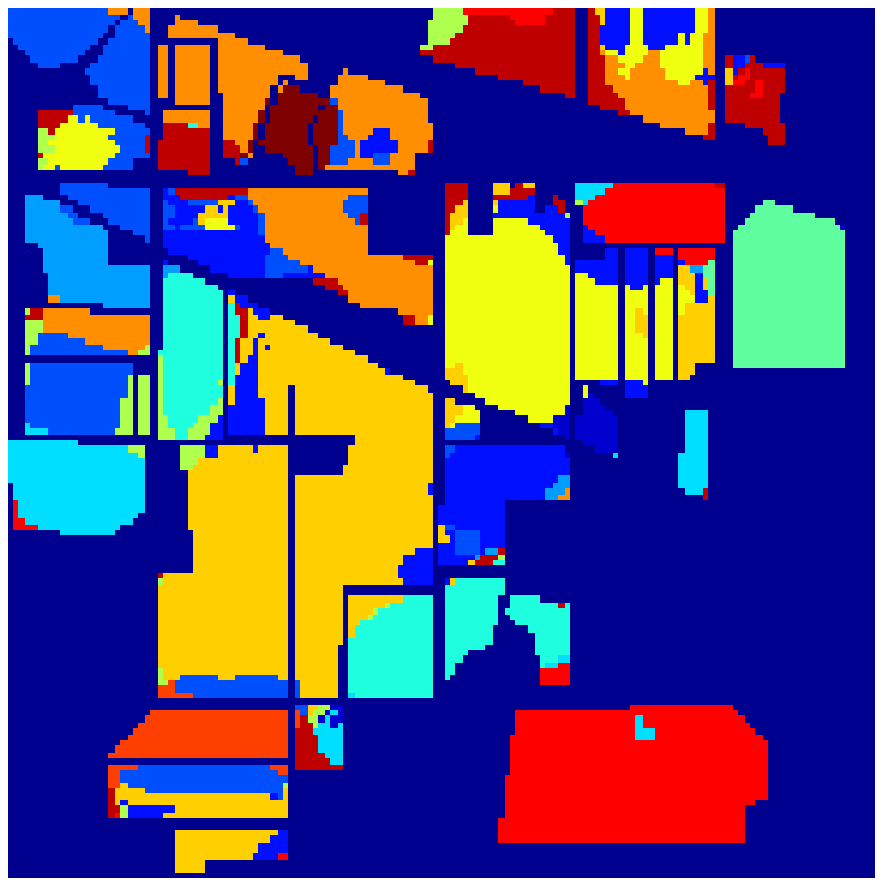}&
\includegraphics[width=0.23\columnwidth]{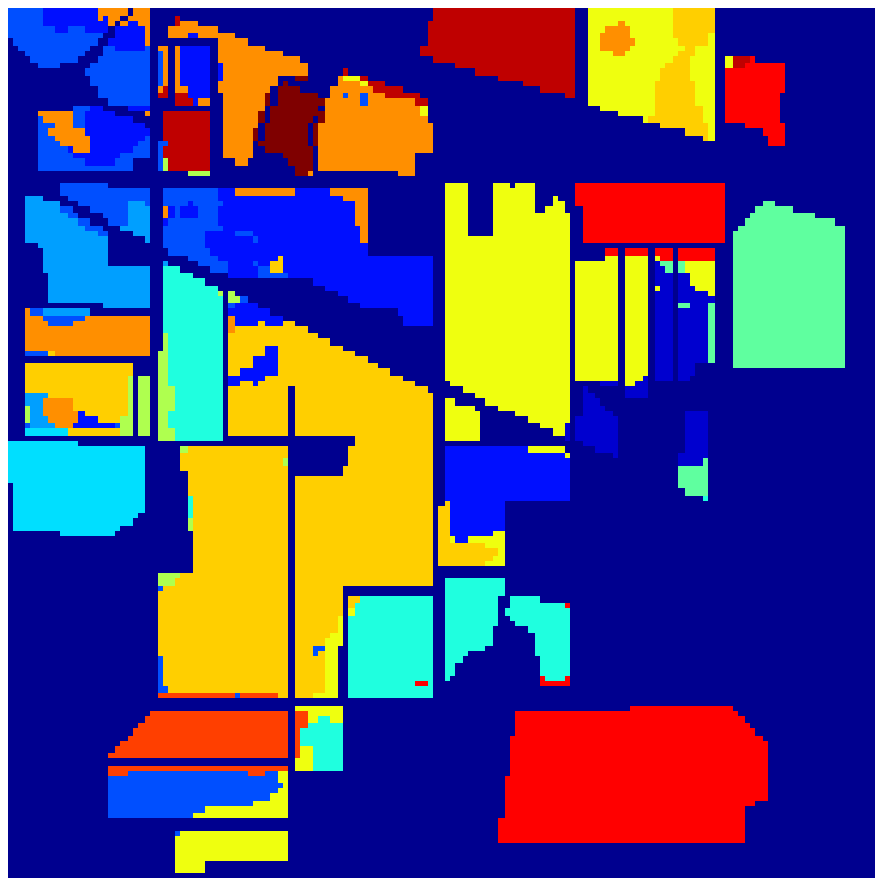}&
\includegraphics[width=0.23\columnwidth]{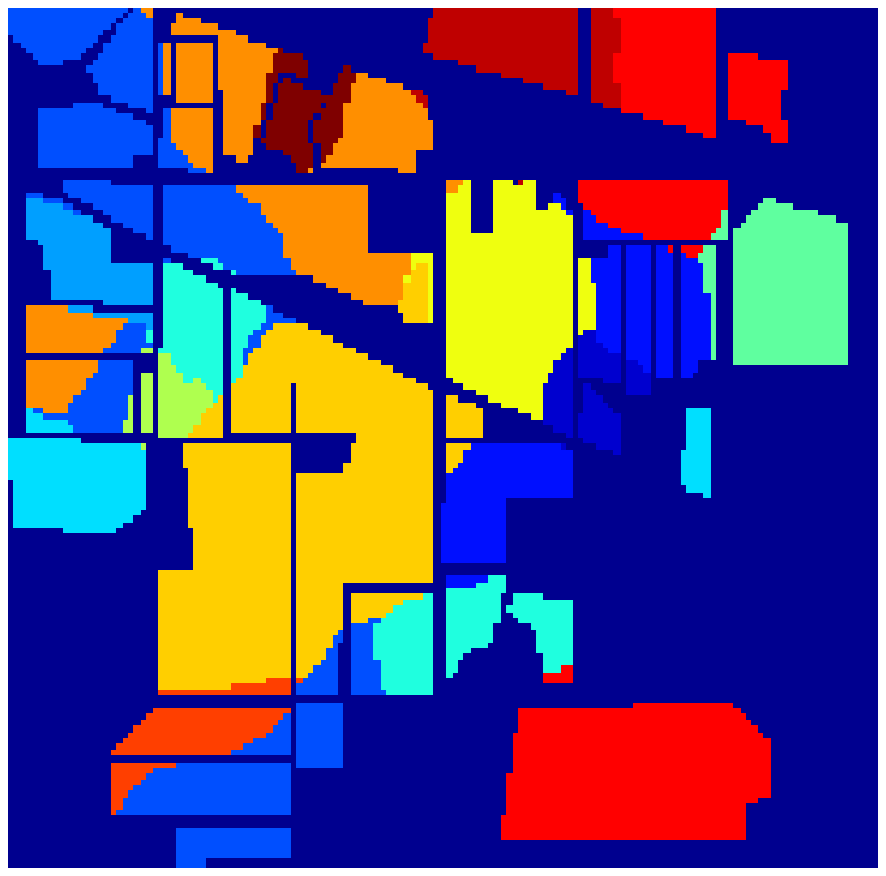}\\
{\small (d)} & {\small (e)} & {\small (f)}
\end{tabular}
\end{center}
\caption{Classification maps for the Indian Pines hyperspectral image using different approaches, with just 5 points per class in the training set. In (a) ground truth, (b) the linear SVM, (c)the estimated RBF SVM, (d) kernel SVM with $KMM$ and $s=15$, (e)  kernel SVM with $KCN$ and $s=7$, (f) kernel SVM with $KCN$ and $s=15$.} \label{fig:result_order}
\end{figure}

\begin{figure}[h]
\begin{center}
\begin{tabular}{ccccc}
\includegraphics[width=0.16\columnwidth]{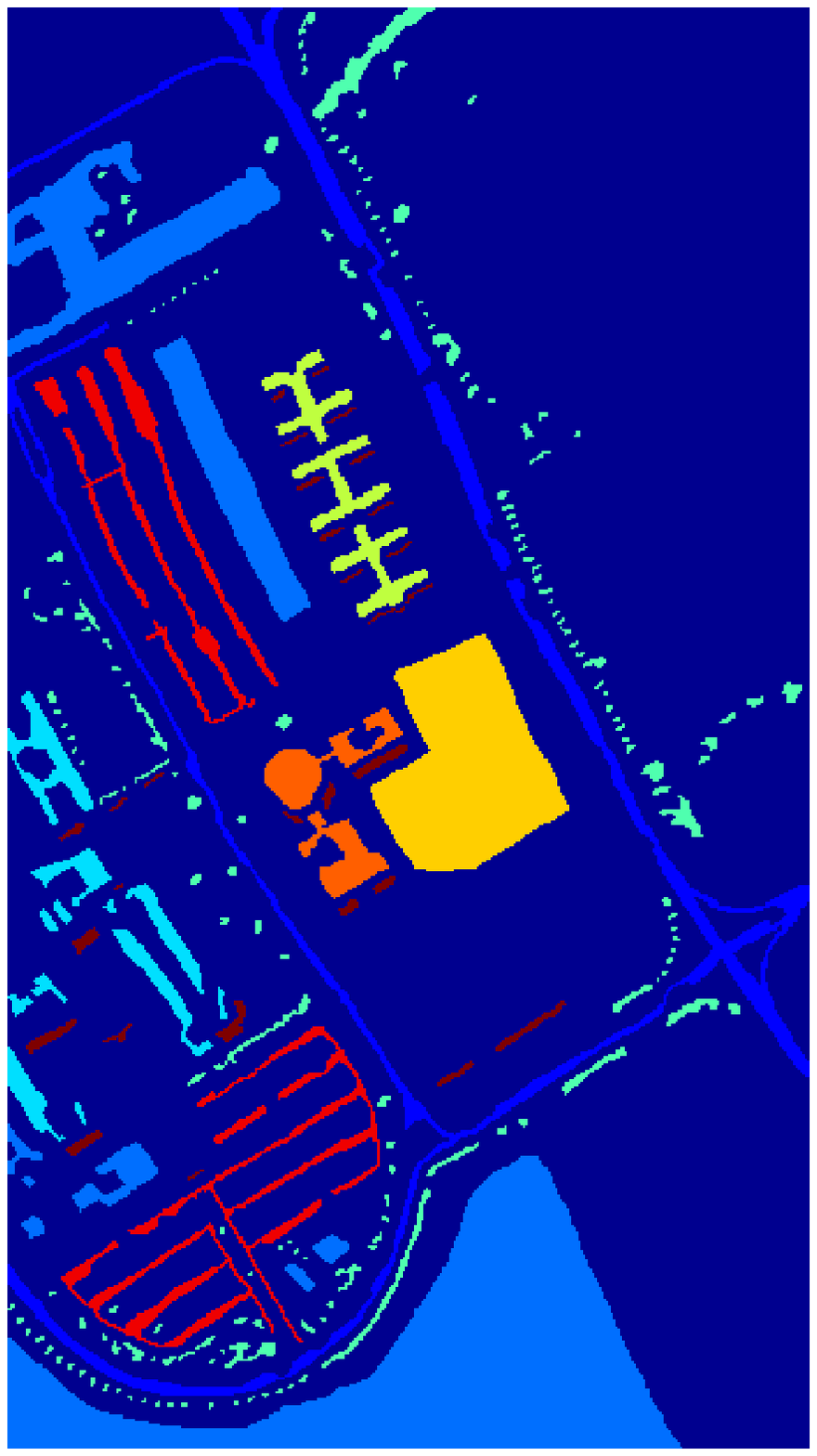}&
\includegraphics[width=0.16\columnwidth]{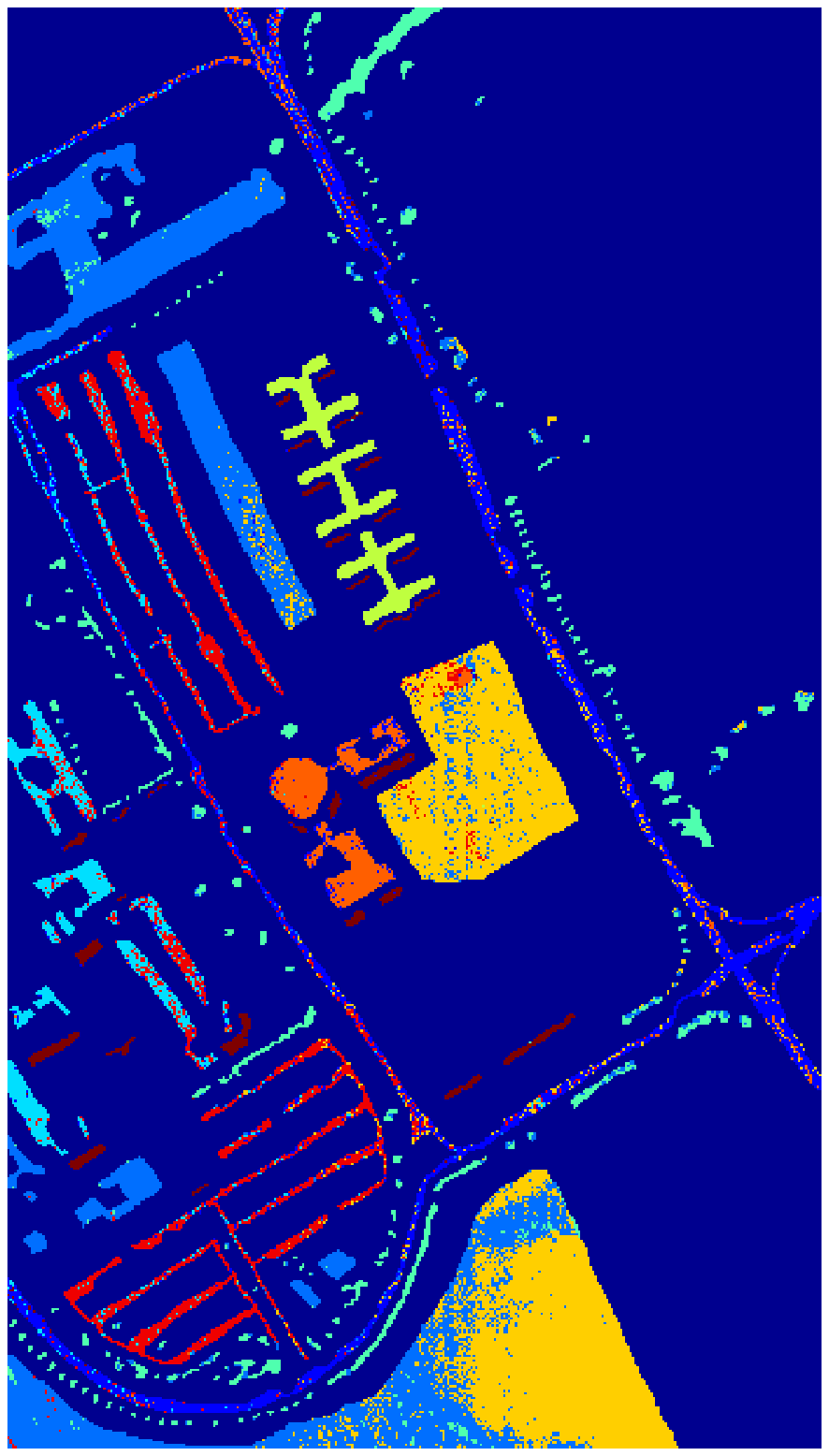}&
\includegraphics[width=0.16\columnwidth]{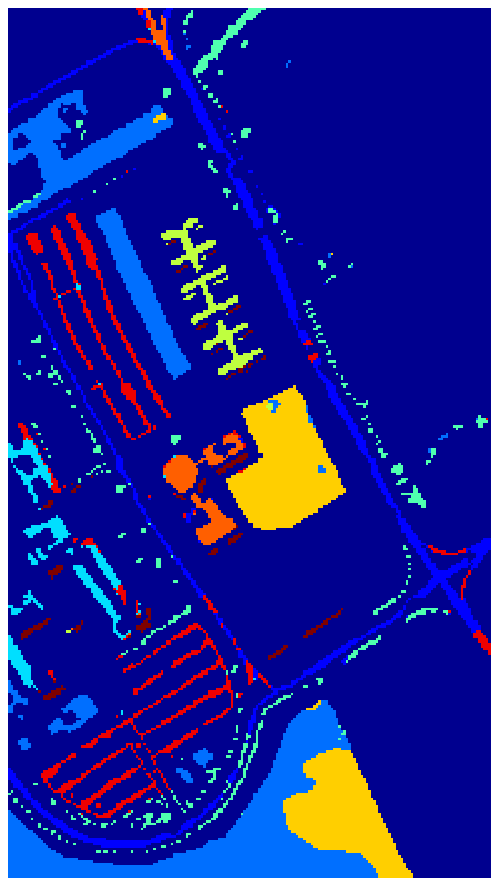}&
\includegraphics[width=0.16\columnwidth]{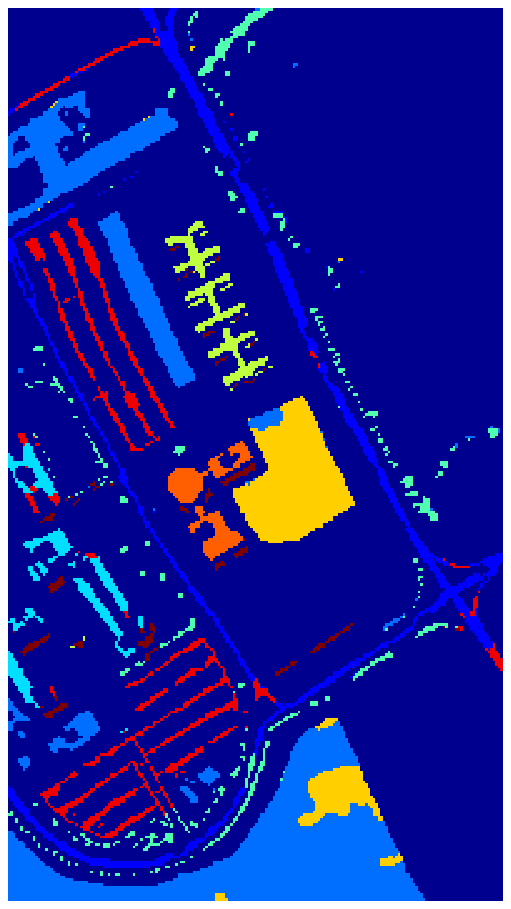}&
\includegraphics[width=0.16\columnwidth]{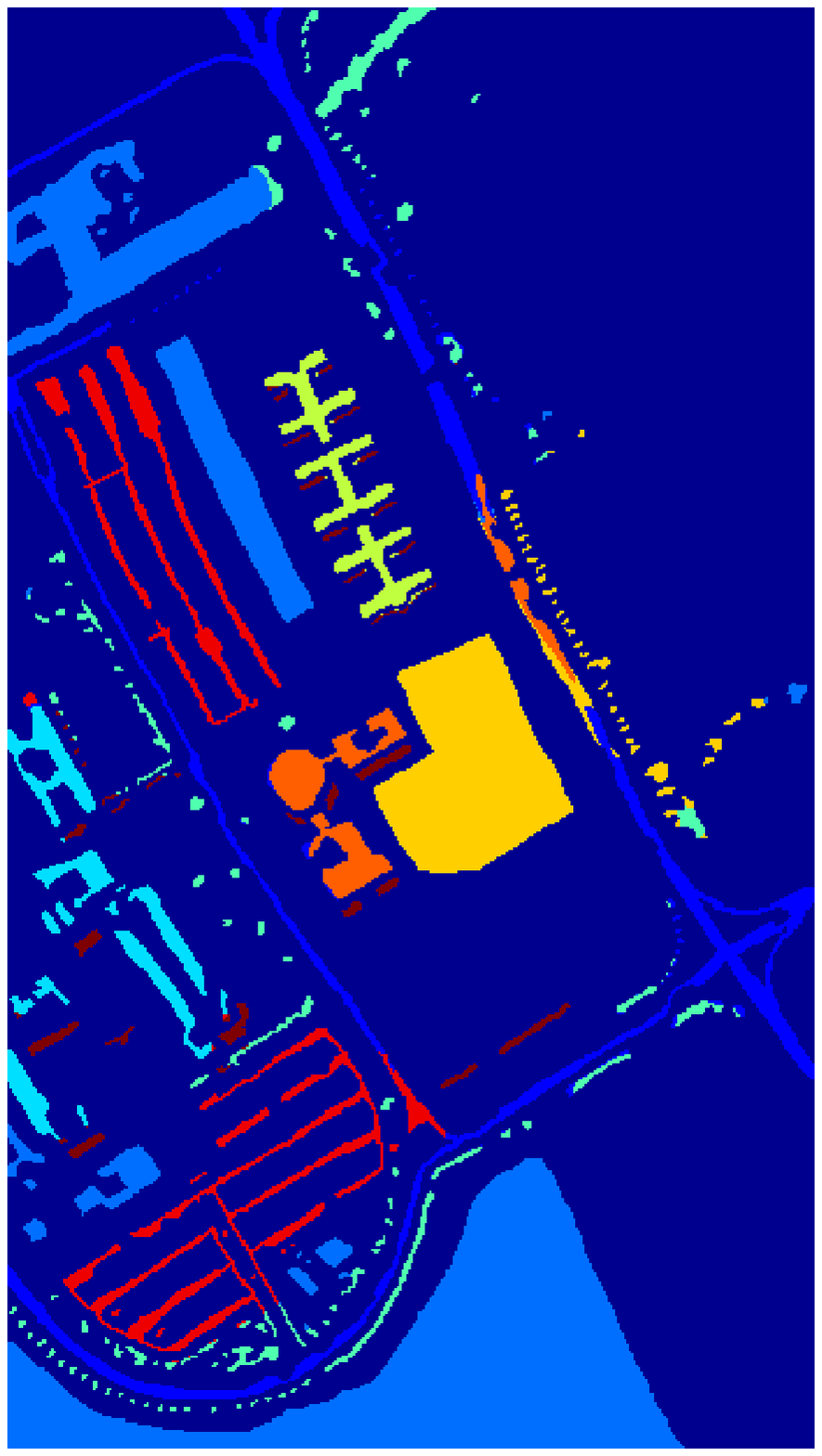}\\
{\small (a)} & {\small (b)} & {\small (c)}&{\small (d)} & {\small (e)} 
\end{tabular}
\end{center}
\caption{Classification maps for the Pavia hyperspectral image using different approaches, with just the classical training set. In (a) ground truth, (b) the linear SVM, (c)the estimated RBF SVM, (d) kernel SVM with $KMM$ and $s=15$, (e)  kernel SVM with $KCN$ and $s=13$.} \label{fig:result_order}
\end{figure}

\section{conclusion}
\label{sec:conclu}

In this article we developed a new method for pixel classification in hyperspectral imaging. The method uses spatial information encoded through distributions of the neighbourhood around each pixel. Even with very simple kernel choices (Gaussian RBF applied directly to raw data) the obtained results are comparable with state-of-the-art. We establish convergence rates for, and prove that we are two-stage consistent. Further improvements are possible by using different feature spaces employing suitable representations of the individual pixels. We believe that we have established an interesting research direction where local distributions are treated as additional features for a supervised learning task, which is of particular interest in hyperspectral imaging where it is difficult to combine spatial and spectral information in a principled way and our approach can be viewed as a step in that direction.

 \clearpage
%\small
%\bibliographystyle{IEEEbib}
%\bibliography{bibliographie}
% \clearpage
%\normalsize

\bibliographystyle{splncs}
%\bibliography{copybibthesis}

\begin{appendices}

\section{Appendix}

\subsection{Morphological Profiles}

Mathematical morphology operators are non-linear image processes based on the spatial structure of the image. Let $f$ be a grey scale image which can be represented by a function. Two basic operators in morphology are the grey-level erosion and the grey-level dilatation whose definition are respectively given by \cite{Jserra}:
\begin{eqnarray}\label{ErosionDilation}
  \varepsilon_{b}(f)(x)& = & \inf_{h\in E} (f(x -h)- b(h)), \\
  \delta_{b}(f)(x)     & = & \sup_{h\in E} (f(x -h)+ b(h)), 
\end{eqnarray}
where $b$ is a structuring function, which introduces the effect of the operators by the geometry of its support as well as the penalizations. We consider for simplicity uniform structuring functions which are formalised by their support set or shape $B$, called structuring element. By concatenation of these two basic morphological operators it is possible to obtain more evolved operators such as the opening and the closing \cite{Jserra} : 

\begin{eqnarray}\label{OpeningClosing}
  \gamma_{B}(f)& = & \delta_{B}\left( \varepsilon_{B}(f) \right), \\
  \varphi_{B}(f) & = & \varepsilon_{B}\left( \delta_{B}(f) \right).
\end{eqnarray}
These  operators  remove  from $f$ all the bright (opening) or dark (closing) structures where the structuring element $B$ cannot fit. However they also modify the value of pixels when $B$ fit. Thus to avoid these artefacts it has been proposed in \cite{Pesaresi} to use geodesic opening and closing. Then by considering a set $\{\gamma_{R}^{(i)} \}$, $i = 1 ... n$, of indexed geodesic openings ,and a set $\{\varphi_{R}^{(i)} \}$, $i = 1 ... n$, of indexed geodesic closings where typically, the index $i$ is associated to the size of the structuring element. Then thanks to the granulometry axiomatic \cite{Jserra} we obtain a scale space representation of an image, which allows an image structures decomposition. Then the Moprhological Profile (MP) of a grey scale image $f$ is defined as $2*n+1$ dimension vector:

\begin{eqnarray}	
MP(x) = \begin{bmatrix}\gamma_{R}^{(1)}(f)(x)\\ \vdots \\ \gamma_{R}^{(n)}(f)(x)\\ f(x)\\ \varphi_{R}^{(1)}(f)(x)\\ \vdots \\ \varphi_{R}^{(n)}(f)(x)\\\end{bmatrix}
\end{eqnarray}
To be able to use the MP on hyperspectral images we first reduce the dimension of the data thanks to PCA, and then project the data on a $d$ dimensional space which is of smaller dimension than the original space. So a hyperspectral image is represented by $d$ grey scale images, then on each of these images we calculate the MP and we concatenate them. Hence, the spatial feature space is of dimension $d*2*n+1$.
\end{appendices}

\subsection{Algorithm for the convolutional kernel mean map}

The formula for the convolutional kernel mean map is defined in equation (\ref{KCN1}) by :
\footnotesize
\begin{eqnarray*}
\widehat{KCN}(x_i,x_j)= \langle\widehat{\mu}_{P(\mathcal{P}_{x_i})},\widehat{\mu}_{P(\mathcal{P}_{x_j})}\rangle  \\
\widehat{KCN}(x_i,x_j)= \sum_{l1 \in \mathcal{P}_{x_i}}  \sum_{l2  \in \mathcal{P}_{x_j}}  \| h(x_{l1}) \|_2 . \| h(x_{l2})\|_2 \\ e^{-\frac{1}{2\beta^2} \| x_{l1}-x_{l2} \|_2 } e^{-\frac{1}{2\sigma^2}.\| \tilde{h}(x_{l1})-\tilde{h}(x_{l2}) \|_2 },
\end{eqnarray*}
\normalsize
 This equation can be rewritten as:
 \begin{eqnarray*}
\widehat{KCN}(x_i,x_j)= \sum_{l1 \in \mathcal{P}_{x_i}}  \sum_{l2  \in \mathcal{P}_{x_j}}  \| h(x_{l1}) \|_2 . \| h(x_{l2})\|_2 \\ e^{-\frac{1}{2} \|  \varpi (x_{l1})-\varpi(x_{l2}) \|_2 },
\end{eqnarray*}
where $x \in \mathbb{R}^2$ corresponds to the position of the pixel, whereas  $h(x) \in \mathbb{R}^D$ is a pixel value and so a spectrum of dimension $D$. For the following formula we write $x^{(1)}$ and $x^{(2)}$ respectively the first and the second spatial coordinate, and $h(x)^{(i)}$ the $i$-th coordinate of the vector $h(x)$.  Finally $\varpi$ is defined by :
\begin{eqnarray}	
\varpi (x) = \begin{bmatrix} \frac{x^{(1)}}{\beta^2}\\ \frac{x^{(2)}}{\beta^2} \\  \frac{h(x)^{(1)}}{\sigma^2} \\ \vdots  \\ \frac{h(x)^{(D)}}{\sigma^2}\\\end{bmatrix}
\end{eqnarray}
then we use the random feature trick on this new vector.
\subsection{Proof of theorems}
\begin{theorem1}
 
 Given that $x \sim P$ an arbitrary probability distribution with variance $\sigma^2$, a Lipschitz continuous function $f :  \mathbb{R} \rightarrow \mathbb{R}$ with constant $C_f$, an arbitrary loss function $\Phi:  \mathbb{R} \rightarrow \mathbb{R}$ that is Lipschitz continuous in the second argument with constant $C_l$ , it follows that :
\begin{eqnarray}
  \mathcal{R}_{\mu}(f) - \mathcal{R}(f)  \leq C_l C_f^2 \mathbb{E}_{(x)}  \| x-\mu_{x} \|^2 \mathbb{E}_{(y)} ( y^2)
\end{eqnarray}

 \end{theorem1}

\begin{proof}
  $\mathcal{R}_{\mu}(f) - \mathcal{R}(f) \leq \mathbb{E}_{(x,y)\sim\mathcal{M}}  \left[  \Phi(f(x).y) -\Phi(f(\mu_{x}).y) \right]$\\
    $\mathcal{R}_{\mu}(f) - \mathcal{R}(f)  \leq \mathbb{E}_{(x,y)\sim\mathcal{M}}  | \Phi(f(x).y) -\Phi(f(\mu_{x}).y) |$
 Since $\Phi$ is Lipschitz continuous we have:
       $\mathcal{R}_{\mu}(f) - \mathcal{R}(f)  \leq C_l \mathbb{E}_{(x,y)\sim\mathcal{M}}  | f(x).y -f(\mu_{x}).y |$.\\
Thanks to the Cauchy-Schwarz inequality we have:\\
$\mathcal{R}_{\mu}(f) - \mathcal{R}(f)  \leq C_l \mathbb{E}_{(x)}  (f(x)-f(\mu_{x}) )^2 \mathbb{E}_{(y)} ( y^2)$
Since $f$ is Lipschitz continuous we have:\\
$\mathcal{R}_{\mu}(f) - \mathcal{R}(f)  \leq C_l C_f^2 \mathbb{E}_{(x)}  \| x-\mu_{x} \|^2 \mathbb{E}_{(y)} ( y^2)$
\end{proof}\\
\begin{theorem3}
 
 Given that $f :  \mathbb{R} \rightarrow \mathbb{R}$ is a Lipschitz continuous function  with constant $C_f$, an arbitrary loss function $\Phi:  \mathbb{R} \rightarrow \mathbb{R}$ that is Lipschitz continuous with constant $C_{l2}$ , it follows that :
\small
\begin{eqnarray*}
  \hat{\mathcal{R}}_{\hat{\mu}}(f) -\hat{\mathcal{R}}_{\mu}(f)  \leq  1/n. C_{l}C_f^2 \hat{\mathbb{E}}\left( \| \mu_x -\hat{\mu}_x \|^2\right)\hat{\mathbb{E}}\left( (y)^2\right)
\end{eqnarray*}
\normalsize
 \end{theorem3}
\begin{proof}
  $\hat{\mathcal{R}}_{\hat{\mu}}(f) -\hat{\mathcal{R}}_{\mu}(f) \leq 1/n. \left[ \sum_{i=1}^n  \Phi(f(\hat{\mu}_{x_i}).y_i) -\Phi(f(\mu_{x_i}).y_i) \right]$\\
    $\hat{\mathcal{R}}_{\hat{\mu}}(f) -\hat{\mathcal{R}}_{\mu}(f) \leq 1/n. \left( \sum_{i=1}^n |  \Phi(f(\hat{\mu}_{x_i}).y_i) -\Phi(f(\mu_{x_i}).y_i) | \right)$
 Since $\Phi$ is Lipschitz continuous we have:
       $\hat{\mathcal{R}}_{\hat{\mu}}(f) -\hat{\mathcal{R}}_{\mu}(f) \leq 1/n. C_{l} \left( \sum_{i=1}^n | ( f(\hat{\mu}_{x_i}) -f(\mu_{x_i})).y_i | \right)$.\\
Thanks to the Cauchy-Schwarz inequality we have:\\
$\hat{\mathcal{R}}_{\hat{\mu}}(f) -\hat{\mathcal{R}}_{\mu}(f) \leq 1/n. C_{l} \left( \sum_{i=1}^n |  y_i |^2 \right) \left( \sum_{i=1}^n |  f(\hat{\mu}_{x_i}) -f(\mu_{x_i}) |^2 \right)$
Since $f$ is Lipschitz continuous we have:\\
$\hat{\mathcal{R}}_{\hat{\mu}}(f) -\hat{\mathcal{R}}_{\mu}(f) \leq 1/n. C_{l}C_f^2 \left( \sum_{i=1}^n (  y_i )^2 \right) \left( \sum_{i=1}^n \|  \hat{\mu}_{x_i} -\mu_{x_i} \| ^2 \right)$
\end{proof}\\
\normalsize

\subsection{Assessing the accuracy of the classification on hyperspectral data}

Most of the time, when one does a classification on hyperspectral data, we do not face a binary classification. Hence we have to handle a more tricky classification. To evaluate this classification we use what is called a confusion matrix. Each column of the matrix represents the number of occurrences of an estimated class, while each row represents the number of occurrences of a real class. If we write $C$ the confusion matrix of a classification then $C_{ij}$ is the number of pixels of class $i$ assign to the class $j$ by the classifier. Let us write $N_c$ the number of class.

\begin{definition}
 The overall accuracy (OA) is the percentage of correctly classified pixels:
\begin{eqnarray*}
\mbox{OA}=\frac{\sum_{i=1}^{N_c} C_{ii}}{\sum_{j=1}^{N_c} \sum_{i=1}^{N_c} C_{ij}}
\end{eqnarray*}
\end{definition}

\begin{definition}
 The average accuracy (OA) is the mean of accuracy of each class for all pixels:
\begin{eqnarray*}
\mbox{AA}=1/N_c.\sum_{i=1}^{N_c} \frac{ C_{ii}}{\sum_{j=1}^{N_c}C_{ij}}
\end{eqnarray*}
\end{definition}

\begin{definition}
 The kappa statistic  is a statistical measure of agreement.
It is the percentage agreement corrected by the level of agreement that could be expected due to chance. Let us define : 
\begin{eqnarray*}
\mbox{Po}=\mbox{OA}
\end{eqnarray*}
\begin{eqnarray*}
\mbox{Pe}=\frac{\sum_{l=1}^{N_c} (\sum_{k=1}^{N_c}C_{lk})(\sum_{k=1}^{N_c}C_{kl})}{\sum_{j=1}^{N_c} \sum_{i=1}^{N_c} C_{ij}}
\end{eqnarray*}

\begin{eqnarray*}
\mbox{kappa statistic}=\frac{ \mbox{Po}-\mbox{Pe}}{1-\mbox{Pe}}
\end{eqnarray*}
\end{definition}
\end{document}